\documentclass{egpubl}
\usepackage{eg2020}

\ShortPresentation
\usepackage[T1]{fontenc}
\usepackage{dfadobe}

\usepackage{cite}  
\BibtexOrBiblatex%

\electronicVersion%
\PrintedOrElectronic%

\ifpdf\usepackage[pdftex]{graphicx}\pdfcompresslevel=9
\else\usepackage[dvips]{graphicx}\fi

\usepackage{egweblnk}

\usepackage{todonotes}
\usepackage{amssymb}
\usepackage{amsmath}
\usepackage{mathtools}
\usepackage{booktabs}
\usepackage{multirow}
\usepackage{subfigure}

\newcommand{\mr}[2]{\multirow{#1}{*}{#2}}
\newcommand{\mc}[3]{\multicolumn{#1}{#2}{#3}}

\def\eg{\emph{e.g.}}   
\def\cf{\emph{cf.}}     
\def\etal{\emph{et~al}.}                   
\def\wrt{w.r.t.}                           

\newcommand{\papertitle}{Adversarial Generation of Continuous Implicit Shape Representations}

\title[\papertitle]{\papertitle}

\author[M. Kleineberg \& M. Fey \& F. Weichert]
{\parbox{\textwidth}{
  \centering
  Marian Kleineberg,
  Matthias Fey and
  Frank Weichert
}
\\
{\parbox{\textwidth}{\centering TU Dortmund University, Germany}}}

\usepackage{capt-of}

\begin{document}

\twocolumn[{
  \renewcommand\twocolumn[1][]{#1}
  \vspace{-1.5cm}
  \maketitle
  \begin{center}
    \centering
    \includegraphics[width=\textwidth]{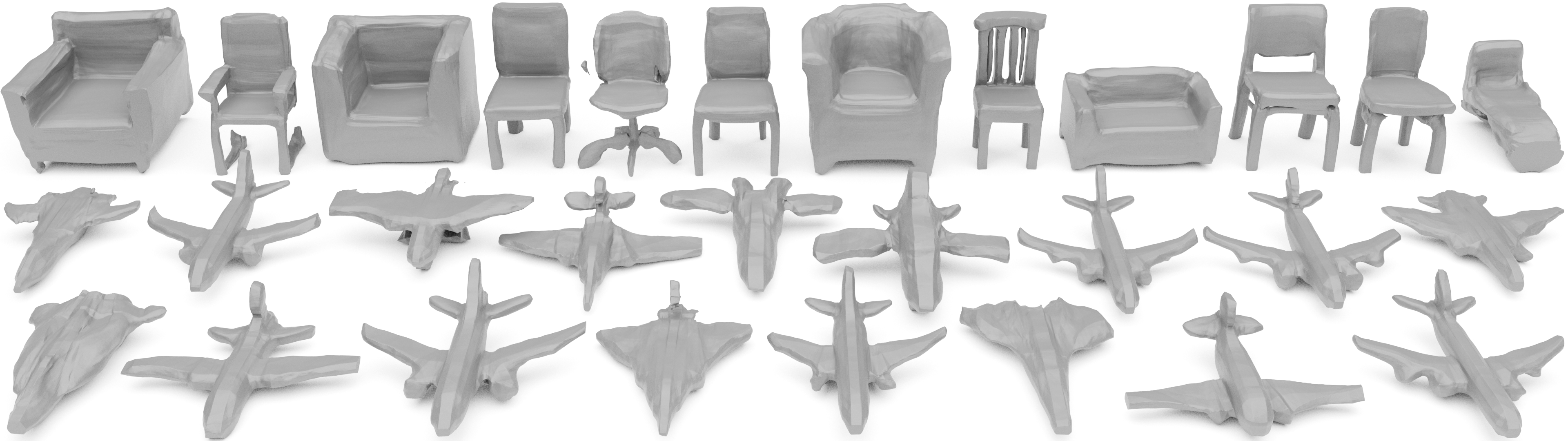}
    \captionof{figure}{Our GAN architecture generates continuous signed distance functions of shapes. Above objects are obtained via Marching Cubes.
    \label{fig:examples}
    }
  \end{center}
  \vspace{1.0cm}
}]

\begin{abstract}
  This work presents a generative adversarial architecture for generating three-dimensional shapes based on signed distance representations.
  While the deep generation of shapes has been mostly tackled by voxel and surface point cloud approaches, our generator learns to approximate the signed distance for any point in space given prior latent information.
  Although structurally similar to generative point cloud approaches, this formulation can be evaluated with arbitrary point density during inference, leading to fine-grained details in generated outputs.
  Furthermore, we study the effects of using either progressively growing voxel- or point-processing networks as discriminators, and propose a refinement scheme to strengthen the generator's capabilities in modeling the zero iso-surface decision boundary of shapes.
  We train our approach on the $\textsc{ShapeNet}$ benchmark dataset and validate, both quantitatively and qualitatively, its performance in generating realistic 3D shapes.
  Our source code is available under \url{https://github.com/marian42/shapegan}.
\end{abstract}



\section{Introduction}%
\label{sec:introduction}

Shape synthesis tackles the task of generating new, diverse, and realistic shapes and is a central problem in many applications requiring high-quality 3D assets.
Inspired by the success of generative adversarial networks (GANs), there has been a tremendous effort in applying those methods in the 3D domain~\cite{atlasnet,3dcnn,pointcloud-gan}.
In contrast to images, however, 3D shapes can be expressed using various different representations~\cite{survey}, each with their own trade-offs across fidelity and efficiency capabilities.

\noindent Recently, the implicit shape representation in the form of signed distance functions (SDFs) emerged to a powerful tool for reconstructing shapes from prior information~\cite{deepsdf}.
Instead of discretizing space, this approach encodes a fully continuous distance field while being both efficient and expressive at the same time.
In practice, this formulation has unlimited resolution, models arbitrary shape topography, ensures watertight surfaces and can be easily converted to voxels, meshes or point clouds.
However, neural networks for continuous shape representations have been exclusively trained to learn latent embeddings via likelihood-based autoencoder or autodecoder schemes~\cite{deepsdf,implicit-fields,occupancy-networks,meta-functions}.

\noindent In this work, we want to explore how we can effectively produce continuous signed distance fields via generative adversarial modeling.
Especially, we study how to combine generated continuous signed distance fields with either voxel- or point-processing discriminators, and show that both schemes can be easily enhanced by the usage of progressively growing GANs~\cite{progressive-gan}.
Furthermore, we propose a refinement strategy to let our point-based discriminator better focus on the zero iso-surface of shapes, which also strengthens the generator's representational power in return.

\begin{figure*}[t]
  \centering
  \subfigure[Voxel-based approach]{
    \includegraphics[height=3.8cm]{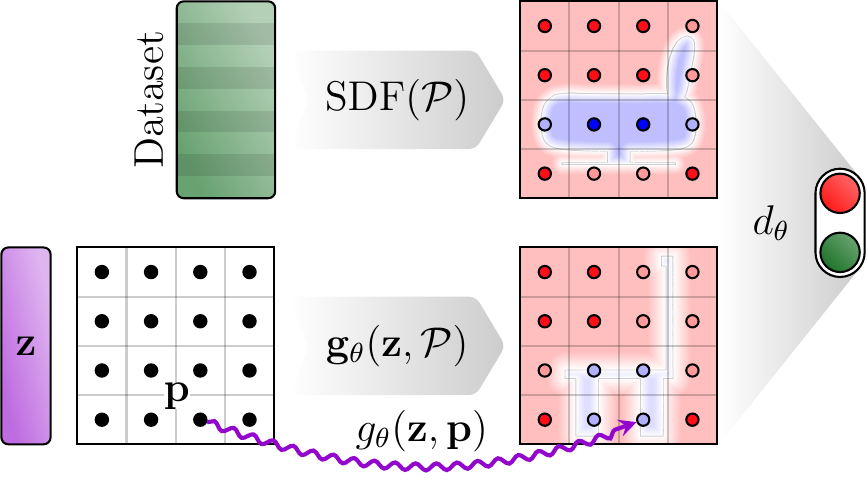}
  }
  \hfill
  \subfigure[Point-based approach]{
    \includegraphics[height=3.8cm]{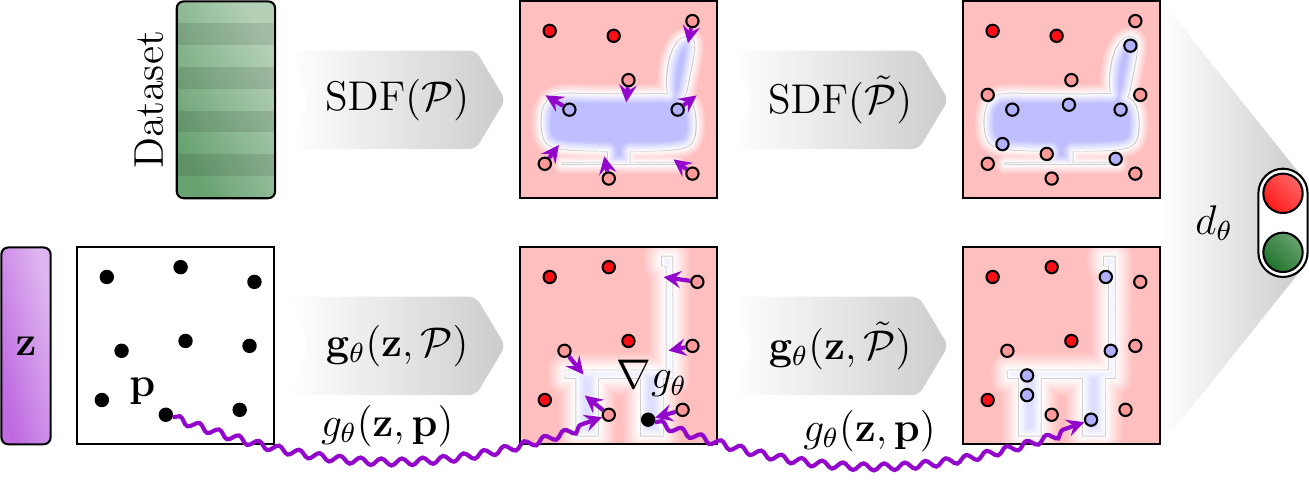}
  }
  \caption{\label{fig:pipeline}Two-dimensional visualization of our generative pipeline: (a) The voxel-based approach uses a fixed number of stationary points as input to a 3D CNN discriminator. In contrast, (b) the point-based approach uniformly samples points in space, samples additional points near the surface based on $\nabla_{\mathbf{p}} \, g_{\theta}(\mathbf{z}, \mathcal{P})$, and inputs the refined point set $\mathcal{\tilde{P}}$ with their generated SDF values into a point-processing discriminator.}
\end{figure*}

\section{Related Work}%
\label{sec:related_work}

3D shape analysis and generation has a long history in computer vision and computer graphics.
Here, we focus on the most directly related works using deep neural networks:

\noindent While neural networks can be used to classify \emph{triangle meshes}~\cite{meshcnn,splinecnn}, they are generally not well-suited to generate them directly due to inherent dicretization problems.
Therefore, the problem of end-to-end mesh generation is typically tackled by learning deformations of primitives~\cite{learning-geometry-images, mesh-deformation}.
However, these approaches are often limited in generating meshes of the same genus as the base mesh.
Further works~\cite{atlasnet, sdm-net} circumvent this problem by assembling multiple deformed primitives, but resulting meshes will not be closed, can have holes and primitives may overlap.

\noindent Successful image generation techniques based on convolutional neural networks (CNNs)~\cite{dcgan} have been also applied to rasterized 3D representations known as \emph{voxel volumes}, where each voxel stores binary occupancy information~\cite{3dcnn} or an SDF value of the voxel center~\cite{3dcnn-sdf}.
However, voxel-based approaches are memory intensive, and are hence limited to low resolutions that can only model coarse-grained structures.
Although data adaptive representations such as octrees~\cite{ogn} can reduce the required memory footprint, it leads to complicated implementations while still being limited to rather small voxel grid sizes~\cite{occupancy-networks}.

\noindent Inspired by unstructured methods operating on unordered point sets~\cite{pointnet}, various methods have been proposed to encode or generate \emph{surface point clouds}~\cite{pointcloud-gan,pointcloud-net,pointcloud-gan-iclr}.
However, those methods rely on reconstructing meshes in a potentially error-prone post-processing step~\cite{psr}, and are generally limited to generating point clouds of fixed sizes.

\noindent Recently, the encoding of shapes has been tackled by learning \emph{continuous implicit functions} in the form of signed distances~\cite{deepsdf} or binary occupancies~\cite{implicit-fields,occupancy-networks,meta-functions}.
These representations have been shown to be both computationally and memory efficient and yet allow for obtaining or visualizing high-resolution geometry via Marching Cubes~\cite{marchingcubes} or direct rendering using Sphere Tracing~\cite{spheretracing}, respectively.
However, neither of these approaches apply generative adversarial networks for implicit shape modeling and instead rely on likelihood-based (encoder-)decoder architectures.
Differentiable variants of Sphere Tracing~\cite{differentiablespheretracing} and Marching Cubes~\cite{deepmarchingcubes} have been proposed to train implicit shape and voxel representations.

\section{Adversarial Generation of Implicit Shape Representations}%
\label{sec:adversarial_generation_of_implicit_shape_representations}

Our generative adversarial network for synthesizing shapes consists of a generator and discriminator.
Given a random latent code, our generator produces a continuous three-dimensional signed distance field, while the discriminator's job is to provide useful feedback to improve the generator.
Our complete pipeline is shown in Figure~\ref{fig:pipeline}, which we will now explain in more detail:

\paragraph*{Signed Distance Functions.}

Given a spatial point $\mathbf{p} \in \mathbb{R}^3$, the \emph{signed distance function} $\mathrm{SDF}(\mathbf{p}) \in \mathbb{R}$ encodes the point's distance to its closest surface point, where the sign indicates whether $\mathbf{p}$ lies inside ($-$) or outside ($+$) the object.
In contrast to binary occupancy information, signed distance functions yield additional useful properties.
For example, it allows us to sample a surface point cloud $\mathcal{S} = \{ \mathbf{s}_1, \ldots, \mathbf{s}_N \}$ of arbitrary cardinality $N$ by translating uniformly sampled points $\mathbf{p}_i \in \mathcal{P} = \{ \mathbf{p}_1, \ldots, \mathbf{p}_N \}$ to their closest surface point:
\begin{equation}
  \label{eq:surface_points}
  \mathbf{s}_i = \mathbf{p}_i - \mathrm{SDF}(\mathbf{p}_i) \cdot \nabla_{\mathbf{p}_i} \, \mathrm{SDF}(\mathbf{p}_i)
\end{equation}

\paragraph*{Generator.}

We model our generator in close analogy to the $\textsc{DeepSDF}$~\cite{deepsdf} autodecoder architecture.
Given a compact, low-dimensional encoding $\mathbf{z} \in \mathbb{R}^d$ of a complete shape, the $\textsc{DeepSDF}$ decoder $g_{\theta}$ learns the mapping
\begin{equation}
  g_{\theta}(\mathbf{z}, \mathbf{p}) \approx \mathrm{SDF}(\mathbf{p}),
\end{equation}
where $g_{\theta}$ is parametrized via trainable parameters $\theta$ and is implemented as an MLP without the usage of any convolutional layers.
Since $g_{\theta}$ is conditioned on a latent vector $\mathbf{z}$, the same neural network can be used to model the SDFs of multiple objects.
Note that $g_{\theta}$ processes only a single point, but can be nonetheless effectively trained against the ground-truth SDF value.
Overall, this formulation is advantageous over signed distance voxel grids~\cite{3dcnn-sdf} since it allows us to model the SDF as a continuous function.
However, it cannot be directly applied to a GAN setup since we argue that a discriminator is not able to distinguish between real and fake solely based on the signed distance of a single sample point.
Instead, the discriminator needs to validate the quality of the signed distance field as a whole.
Therefore, we provide context to the discriminator by evaluating the generator for a batch of points $\mathcal{P}$, and inject the latent information $\mathbf{z}$ to each point $\mathbf{p}_i \in \mathcal{P}$ individually:
\begin{equation}
  {\mathbf{g}_{\theta}(\mathbf{z}, \mathcal{P})}_i = g_{\theta}(\mathbf{z}, \mathbf{p}_i)
\end{equation}
This scheme shares similar advantages to the $\textsc{DeepSDF}$ decoder.
Although it requires fixed point sizes during training (in analogy to related point cloud approaches), we can still obtain high-resolution meshes during inference by varying the sampling density.

\noindent We now proceed to present two discriminator variants, a \emph{voxel-based} and a \emph{point-based} approach, to discriminate between real and generated signed distance fields:

\paragraph*{Voxel-based Discriminator.}

In our first variant, we explore the possibility of using a 3D CNN as our discriminator $d_{\theta}$ similar to the one proposed by Wu~\etal~\cite{3dcnn}.
Here, both $\mathbf{g}_{\theta}(\mathbf{z}, \mathcal{P})$ and $d_{\theta}(\mathbf{g}_{\theta}(\mathbf{z}, \mathcal{P}), \mathcal{P})$ expect input points $\mathcal{P}$ to describe a regular grid of fixed resolution.
Hence, $\mathbf{g}_{\theta}$ and $d_{\theta}$ can be easily combined by rearranging the generated SDF values from $\mathbf{g}_{\theta}$ into a voxel volume.
Although being straightforward to implement, this approach shares \emph{some} of the disadvantages of related voxel-based approaches:
It only makes use of a fixed number of stationary points to discriminate between real and fake examples, and is furthermore quite memory-inefficient to train.
However, since $g_{\theta}$ is modeled to be continuous (in contrast to related approaches~\cite{3dcnn,3dcnn-sdf}), we can even expect reasonable outputs for points never seen during training, \cf~Section~\ref{sec:experiments}.

\paragraph*{Point-based Discriminator.}

An alternative to modeling the discriminator $d_{\theta}$ in a voxel-based fashion builts upon recent advances in permutation-invariant point-processing networks.
Instead of only providing the discriminator with stationary point samples, we may input \emph{any} point by making use of the $\textsc{PointNet}$ architecture~\cite{pointnet} as our discriminator $d_{\theta}$~\cite{pointcloud-gan,pointcloud-gan-iclr}.
Here, uniformly distributed points are first transformed \emph{independently} into a high-dimensional space before a joint representation is obtained using the permutation-invariant max-pooling operation
\begin{equation}
  d_{\theta}(\mathbf{g}_{\theta}(\mathbf{z}, \mathcal{P}), \mathcal{P}) = \gamma_{\theta} \left( \max_{\mathbf{p} \in \mathcal{P}} h_{\theta}(g_{\theta}(\mathbf{z}, \mathbf{p}), \mathbf{p}) \right) \in \mathbb{R}
\end{equation}
with $\mathbf{p} \sim {\mathcal{U}(-1, 1)}^{3}$, and $\gamma_{\theta}$ and $h_{\theta}$ denoting trainable MLPs~\cite{pointnet}.
In contrast to the original $\textsc{PointNet}$ proposal, we also inject the respective signed distance to each raw point so that $g_{\theta}$ takes both positional and signed distance information into account.
Compared to the voxel-based approach, this formulation allows $g_{\theta}$ to know about \emph{any} point in space, instead of solely being required to generate reasonable outputs for a fixed number of stationary points.

\noindent Note that $d_{\theta}$ is not limited to the $\textsc{PointNet}$ architecture, but may employ any network architecture that is able to process irregularly structured data in a permutation-invariant fashion.
For example, the recent works in the field of \emph{geometric deep learning} provides a large number of operators to choose from~\cite{survey}, with potential capabililties to also take relational information into account.
Extensions like $\textsc{PointNet++}$~\cite{pointnet2} may even improve upon the results presented in this paper since those methods should be able to capture more fine-grained details.
We leave the usage of those discriminators for future work.

\paragraph*{Zero Iso-surface Decision Boundary.}

For reconstructing meshes from SDFs, we are mostly interested in the precise modeling of the zero iso-surface decision boundary and care less about maintaining a metric SDF for larger distances.
While $\textsc{DeepSDF}$ overcomes this problem via a more aggressively sampling near the surface of an object, this solution cannot be applied directly since we do not know about the object's surface in advance.
Nonetheless, $\mathbf{g}_{\theta}(\mathbf{z}, \mathcal{P})$ already provides great capabililties to accurately predict the coarse-grained surface of an object.
Following up on Equation~\eqref{eq:surface_points}, we strengthen the generator's representational power by sampling additional points near the surface of an object based on the \emph{generated} SDF values of uniformly distributed points
\begin{equation}
  \mathbf{\tilde{g}}_{\theta}(\mathbf{z}, \mathcal{P}) = \mathbf{g}_{\theta}(\mathbf{z}, \mathcal{P}) \,\,\bigcup\limits_{\mathclap{\substack{\mathbf{p} \in \mathcal{P}\\|g_{\theta}(\mathbf{z}, \mathbf{p})| < \delta}}}\,\,\big\{ g_{\theta}(\mathbf{z}, \mathbf{p} - g_{\theta}(\mathbf{z}, \mathbf{p}) \cdot \nabla_{\mathbf{p}} \, g_{\theta}(\mathbf{z}, \mathbf{p}) + \epsilon) \big\}
\end{equation}
with $\epsilon \sim {\mathcal{N}(0, \sigma^2)}^3$.
Here, for each point $\mathbf{p} \in \mathcal{P}$ that is sufficiently close to a predicted surface ($|g_{\theta}(\mathbf{z}, \mathbf{p})|<\delta$), we project it onto the surface using the gradients of $g_{\theta}$ \wrt~$\mathbf{p}$, and sample additional points following a gaussian distribution.
The discriminator then takes the \emph{refined point set}
\begin{equation}
  \label{eq:refined_point_set}
  \mathcal{\tilde{P}} = \mathcal{P} \cup \{ \mathbf{p} - g_{\theta}(\mathbf{z}, \mathbf{p}) \cdot \nabla_{\mathbf{p}} \, g_{\theta}(\mathbf{z}, \mathbf{p}) + \epsilon \}
\end{equation}
and their respective SDF values as input, and can hence more specifically draw its attention to the modeling of the zero iso-surface decision boundary.
Note that $\mathbf{\tilde{g}}_{\theta}$ is still fully differentiable and can hence be trained in an end-to-end fashion using SGD.\@

\begin{figure*}[t]
  \centering
  \includegraphics[width=\linewidth]{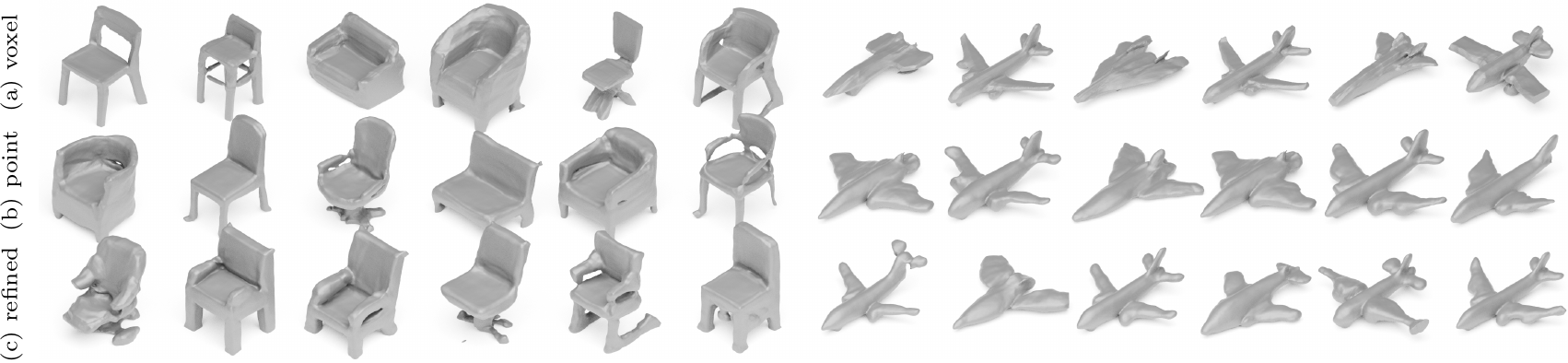}
  \caption{\label{fig:generated_shapes}
    Qualitative examples of generated shapes for (a) the voxel-based discriminator (top), (b) the point-based discriminator (middle) and (c) the refined point-based discriminator (bottom).
  }
\end{figure*}

\begin{figure}[t]
  \centering
  \subfigure[]{
    \includegraphics[width=0.3\linewidth]{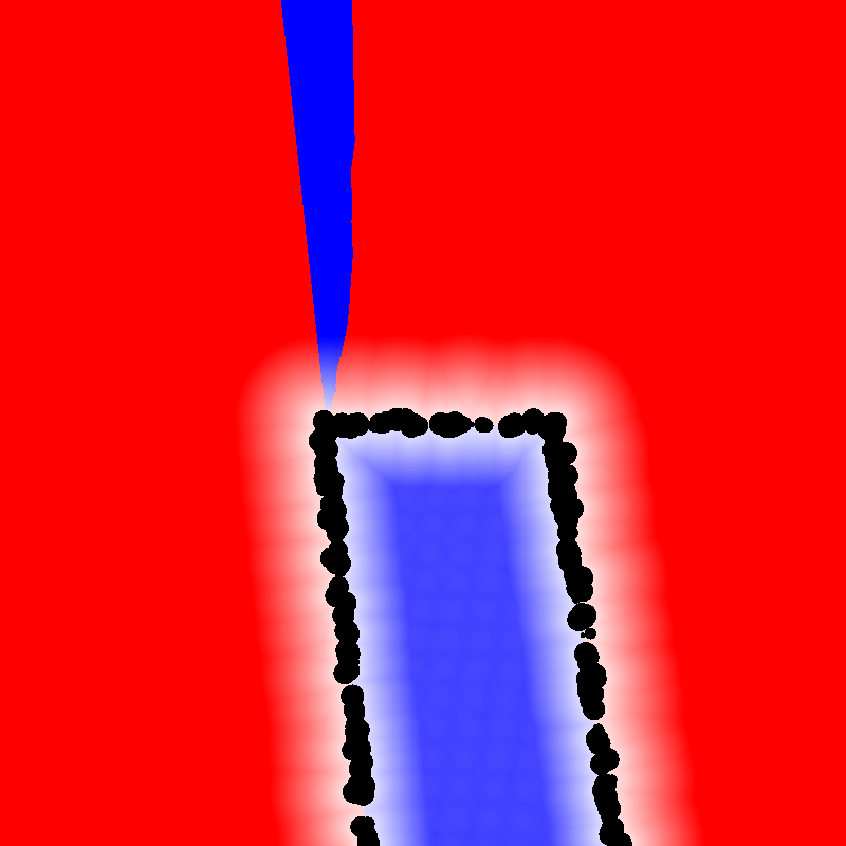}
  }
  \hfill
  \subfigure[]{
    \includegraphics[width=0.3\linewidth]{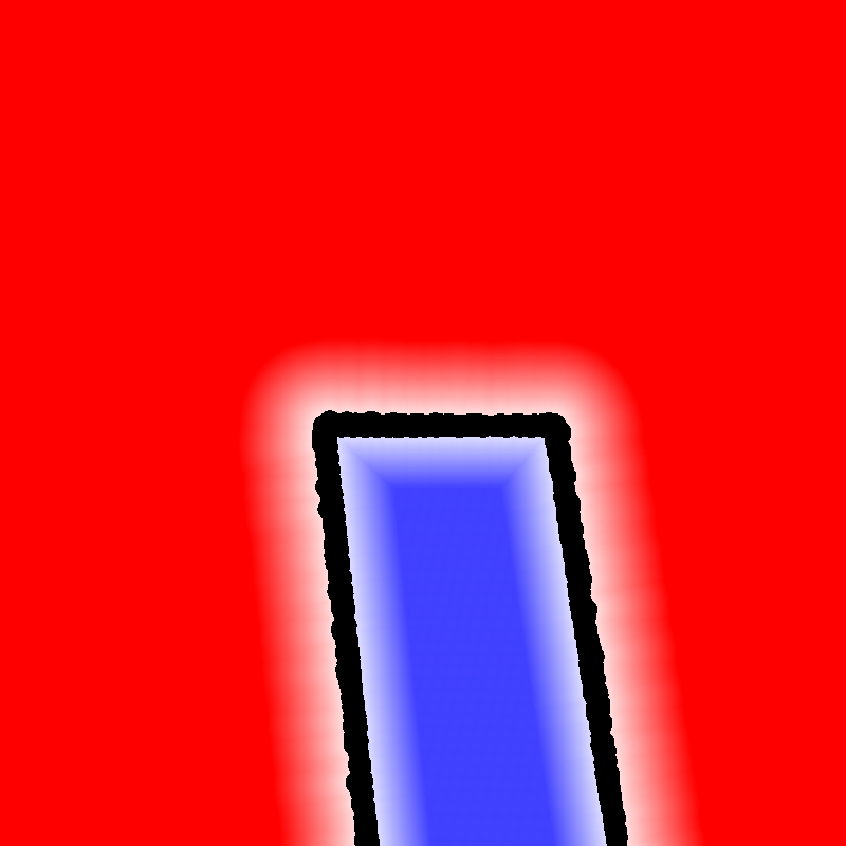}
  }
  \hfill
  \subfigure[]{
    \includegraphics[width=0.3\linewidth]{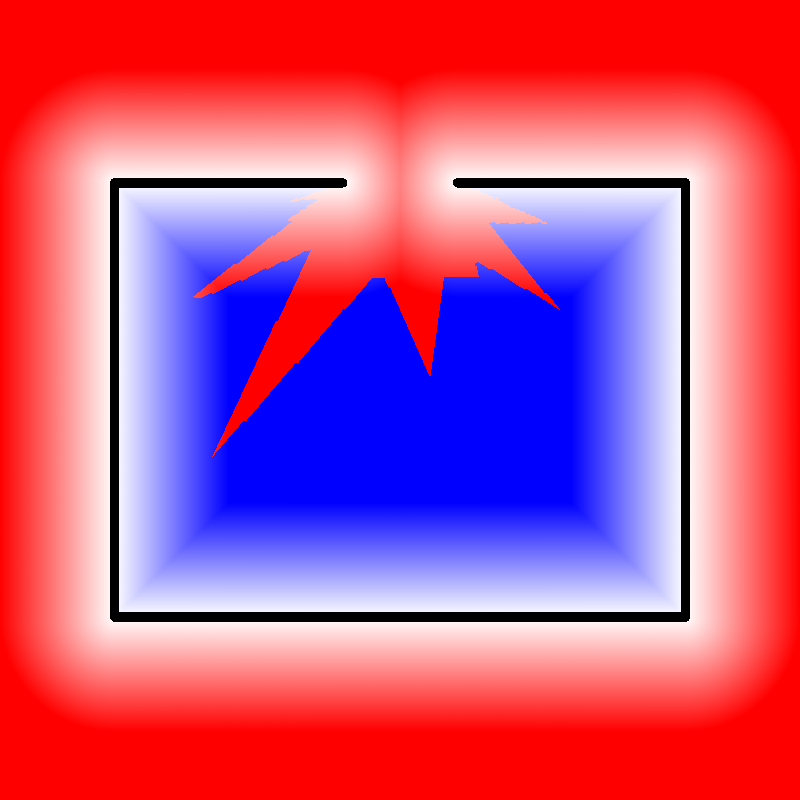}
  }
  \caption{\label{fig:dataprep}
    Calculating SDFs:
    (a) Small triangles with deviating normals may produce artifacts in the SDFs~\cite{deepsdf} and (b) are avoided by using depth information to determine signs.
    (c) Non-watertight meshes lead to discontinuous SDFs and are discarded.
  }
\end{figure}

\begin{figure}[t]
	\centering
	\includegraphics[width=0.19\linewidth]{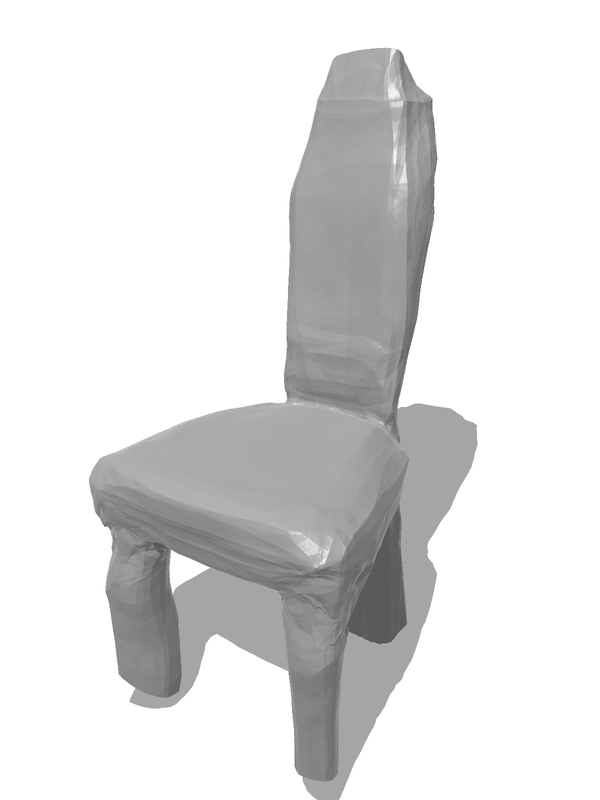}
	\includegraphics[width=0.19\linewidth]{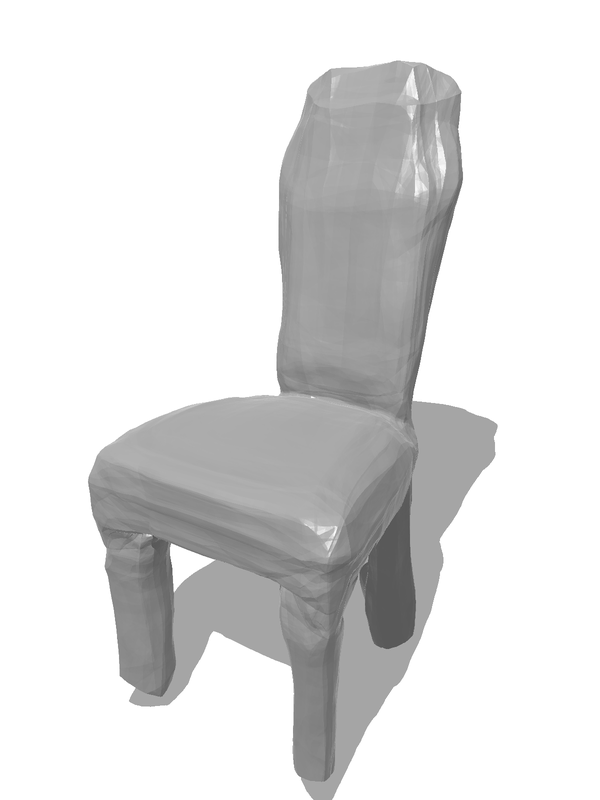}
	\includegraphics[width=0.19\linewidth]{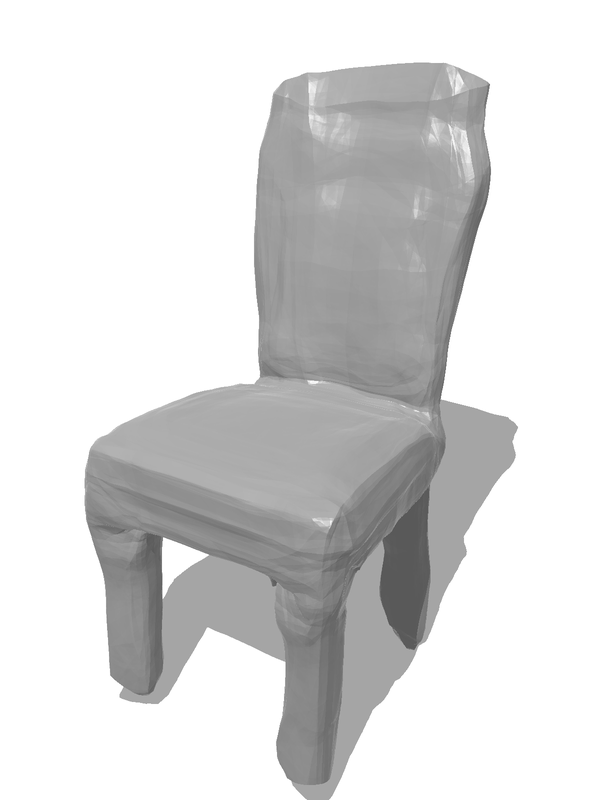}
	\includegraphics[width=0.19\linewidth]{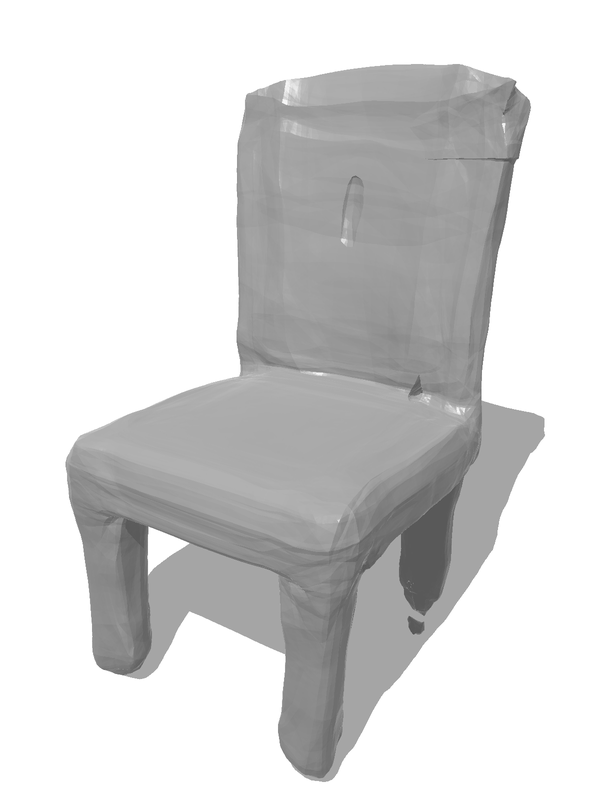}
	\includegraphics[width=0.19\linewidth]{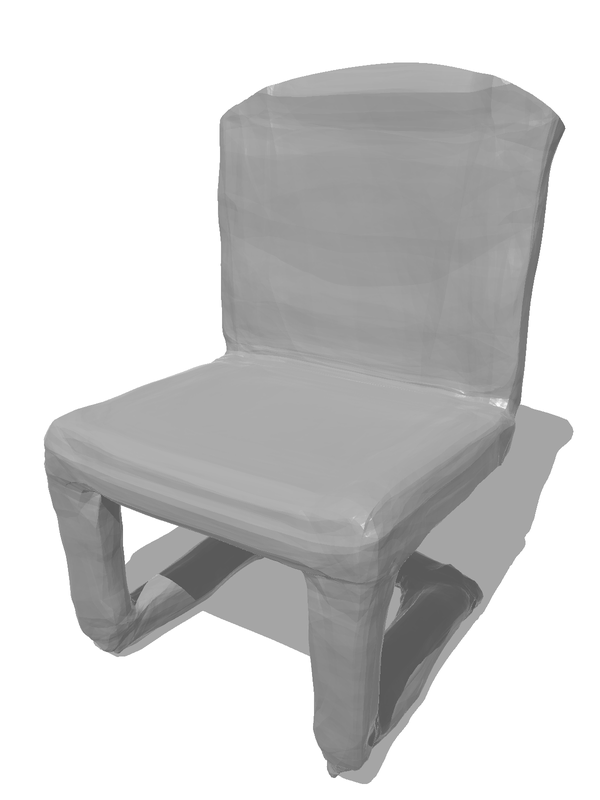}
	\caption{\label{fig:interpolation}
		Latent space interpolation.
		The first and last latent codes were obtained by overfitting shapes of the dataset and the remaining latent codes were obtained by linear interpolation.
		The images are rendered with Sphere Tracing \cite{spheretracing} from signed distance fields given by our generator network.
	}
\end{figure}

\paragraph*{Training.}

For training our generative adversarial networks, we make use of pre-processed ground-truth distance fields $\mathrm{SDF}(\mathcal{P})$ stemming from human designed meshes.
We input both generated and real samples $d_{\theta}(\mathbf{g}_{\theta}(\mathbf{z}, \mathcal{P}), \mathcal{P})$ and $d_{\theta}(\mathrm{SDF}(\mathcal{P}), \mathcal{P})$ to the discriminator $d_{\theta}$, respectively, while optimizing for the typical GAN minimax objective~\cite{gan}.
Furthermore, we noticed that making use of progressively growing GANs~\cite{progressive-gan} improves the training stability and quality of generated objects.
That is, we progressively increase voxel resolution and network depth in our voxel-based discriminator, and simply increase the number of point samples in our point-based architecture as training proceeds.
Our generative network $\mathbf{g}_{\theta}$ does not need to be increased in both cases, since it is already invariant towards specific resolutions by design.

\section{Experiments}%
\label{sec:experiments}

We trained our proposed generative architectures on the chair and airplane categories taken from the $\textsc{ShapeNet}$ repository~\cite{shapenet}.
We randomly divide shapes into train/validation/test sets using an 85\%-5\%-10\% split ratio.
The data preparation pipeline needs to handle non-watertight meshes and meshes with inconsistently oriented normals, as both appear in the dataset, \cf~Figure~\ref{fig:dataprep}.
Our approach to data preparation is based on the method employed by Park \etal~\cite{deepsdf}.
We render each mesh from 50 equidistant camera angles and project the depth buffers back into object space, resulting in a surface point cloud.
For each query point, we then calculate the distance to its closest surface point.
To determine the sign, we transform the sample point into the viewport coordinates of each render.
We use the depth buffers to check if the point is closer to the camera than the surface of the shape at the corresponding pixel.
A point is considered outside the mesh if it is seen by any of the virtual cameras.
We discard objects with discontinuous SDFs and those with less than $1\%$ of uniformly sampled points inside the shape, resulting in $4\,189$ and $2\,156$ examples for the chair and airplane categories, respectively.
Our pre-processing pipeline is publicly available on $\textsc{GitHub}$.\footnote{\url{https://github.com/marian42/mesh\_to\_sdf}}

\paragraph*{Architecture and Parameters.}

Our generator $g_{\theta}$ follows the design of $\textsc{DeepSDF}$~\cite{deepsdf} and is composed of 8 fully connected layers with hidden dimensionalities of $256$, layer normalization and ReLU activiation.
The input is once again reinjected via concatenation after the fourth layer.
We use 128-dimensional latent vectors $\mathbf{z}$ sampled from a normal distribution.
\noindent The voxel-based discriminator uses 3D convolutions with a stride of 2 and a kernel size of 4 to achieve powers of 2 for all intermediate voxel resolutions, followed by two dense layers (128, 1)~\cite{3dcnn}.
We use four steps of progressive growing~\cite{progressive-gan} with samples of increasing resolution ($8^3$, $16^3$, $32^3$, $64^3$).
\noindent The $\textsc{PointNet}$ discriminator~\cite{pointnet} uses 4 layers with weights shared across the point dimension (64, 128, 256 and 512), followed by global max-pooling and 3 dense layers (256, 128, 1).
We uniformly sample $64^3$ points for ground-truth SDF values per object, while only using a small but progressively growing subset of points as input.
For training with refined points $\mathcal{\tilde{P}}$, we sample additional ground-truth values according to Equation~\eqref{eq:refined_point_set}.

\noindent For optimization, we make use of the $\textsc{WGAN-GP}$~\cite{wgan-gp} objective and employ $\textsc{RMSprop}$ with a fixed learning rate of $10^{-4}$.
All models have been trained for a maximum of $2\,000$ epochs, while we select the final model based on validation results.

\begin{table}[t]
  \centering
  \caption{Quantitative evaluation across $\textsc{ShapeNet}$ categories.}\label{tab:results}
  \setlength{\tabcolsep}{2pt}
  \renewcommand{\arraystretch}{0.9}
  \resizebox{\linewidth}{!}{%
  \begin{tabular}{lllccccc}
    \toprule
      & \mc{2}{l}{\textbf{Method}} & \textbf{JSD} & \textbf{MMD-CD} & \textbf{MMD-EMD} & \textbf{COV-CD} & \textbf{COV-EMD} \\
    \midrule
      \mr{5}{\rotatebox{90}{Chair}}
      & \mc{2}{l}{\cite{pointcloud-gan}}      & 0.238          & \textbf{0.0029} & 0.136          & \textbf{33} & 13          \\
      & \mc{2}{l}{\cite{pointcloud-gan-iclr}} & 0.100          & \textbf{0.0029} & 0.097          & 30          & 26          \\
      & \mr{3}{\textbf{Ours}} & voxel         & \textbf{0.076} & 0.0037          & 0.087          & \textbf{33} & \textbf{36} \\
      &                       & point         & 0.082          & 0.0036          & 0.088          & 28          & 29          \\
      &                       & refined       & 0.078          & 0.0037          & \textbf{0.086} & 30          & 28          \\
    \midrule
      \mr{5}{\rotatebox{90}{Airplane}}
      & \mc{2}{l}{\cite{pointcloud-gan}}      & 0.182          & 0.0009          & 0.094          & 31          & 9           \\
      & \mc{2}{l}{\cite{pointcloud-gan-iclr}} & 0.083          & \textbf{0.0008} & 0.071          & 31          & 14          \\
      & \mr{3}{\textbf{Ours}} & voxel         & 0.093          & 0.0019          & \textbf{0.066} & 31          & \textbf{34} \\
      &                       & point         & 0.114          & 0.0026          & 0.078          & 28          & 31          \\
      &                       & refined       & \textbf{0.072} & 0.0019          & 0.070          & \textbf{36} & 31          \\
    \bottomrule
  \end{tabular}}
\end{table}

\paragraph*{Results.}

Qualitative results of our models are shown in Figures~\ref{fig:generated_shapes} and \ref{fig:examples}.
From a visual standpoint, all models generate a diverse set of convincing shapes, and are able to produce, \eg, thin chair legs and a precise modeling of the airplane tail.
In Figure~\ref{fig:interpolation}, we demonstrate latent space continuity of our generator by linearly interpolating between latent codes of dataset shapes.

\noindent For a quantitative evaluation, we follow the experimental protocol of recent generative point cloud approaches~\cite{pointcloud-gan,pointcloud-gan-iclr}.
We report the Jensen-Shannon divergence (JSD) between marginal distributions in the 3D space, and the coverage (COV) and minimum matching distance (MMD) based on the Chamfer distance (CD) and the earth mover's distance (EMD) between point sets, \cf~Table~\ref{tab:results}.
Point clouds are obtained by sampling $2\,048$ points on meshes generated via Marching Cubes.
Given that the Chamfer distance is known to be unreliable~\cite{pointcloud-gan,pointcloud-gan-iclr}, our results achieve better or equal values for the metrics under consideration.
In general, the refined point-based approach improves the results of using uniformly sampled points, while there is not yet a clear winner between the voxel- and point-based approaches.

\begin{figure}[t]
  \centering
  \subfigure[$8^3$]{
    \includegraphics[width=0.29\columnwidth]{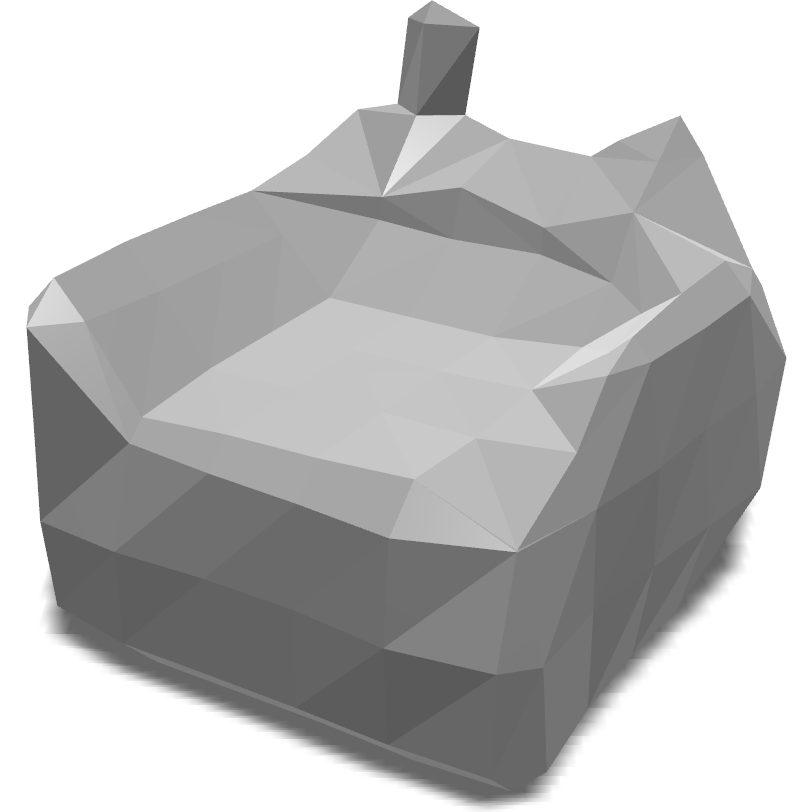}
  }
  \subfigure[$8^3$ interpolated to $128^3$]{
    \hspace{0.1cm}
    \includegraphics[width=0.29\columnwidth]{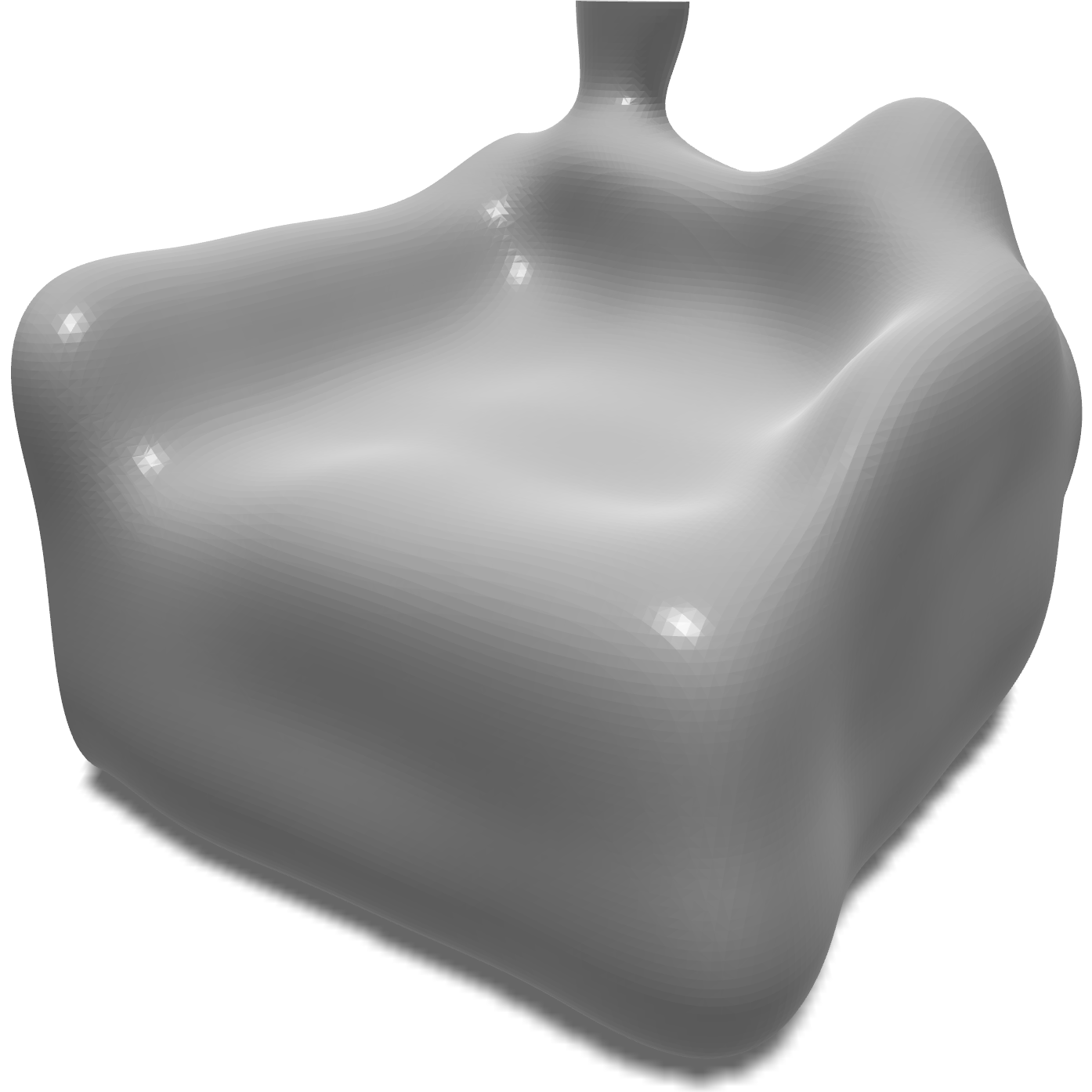}
    \hspace{0.1cm}
  }
  \subfigure[$128^3$]{
    \includegraphics[width=0.29\columnwidth]{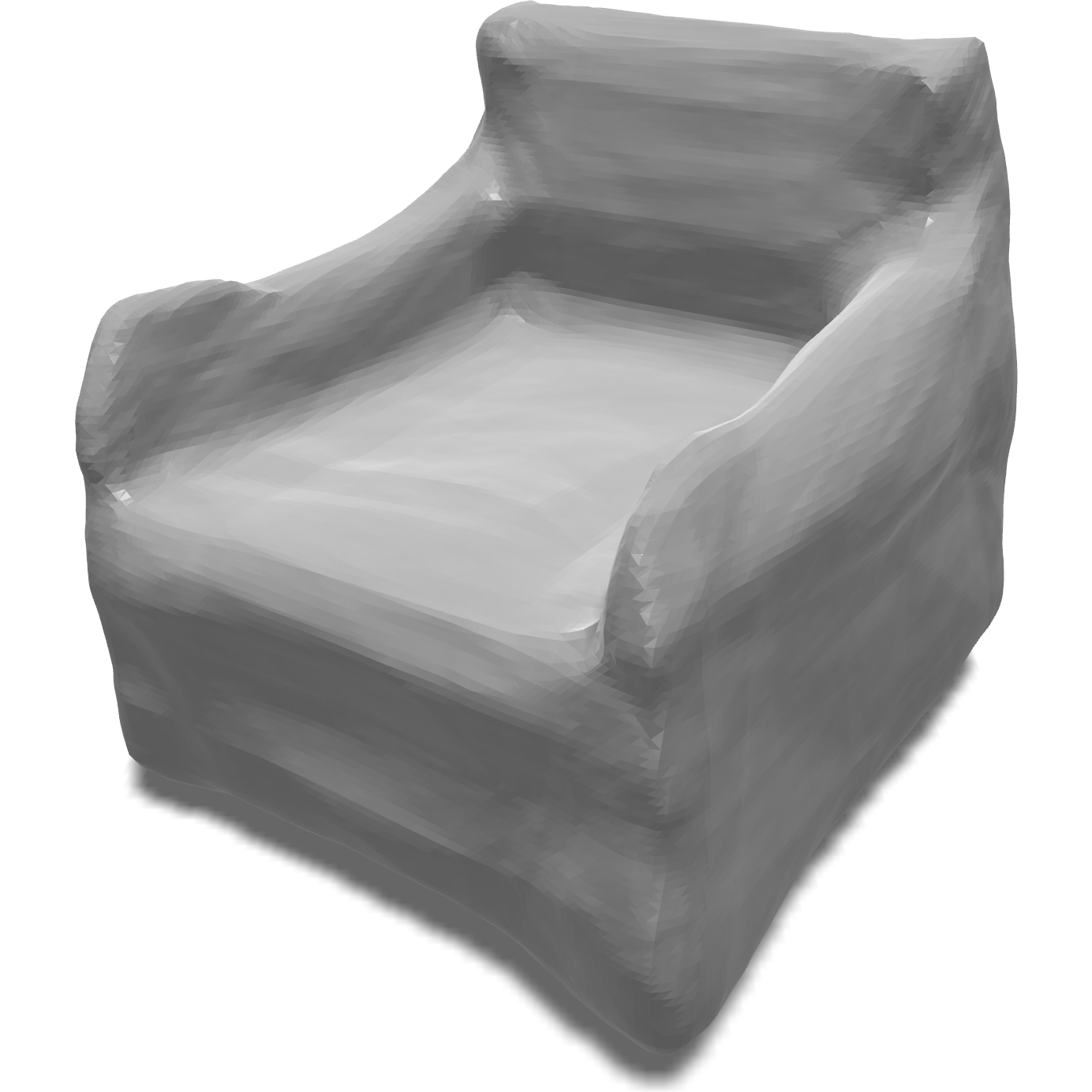}
  }
  \caption{\label{fig:upscaling}
    Generalization capabililties of our voxel-based approach when trained on $8^3$ voxels:
    (a) A random latent code evaluated at $8^3$ raster points,
    (b) $8^3$ raster points linearly upscaled to $128^3$,
    (c) the same latent code evaluated with $128^3$ raster points.
  }
\end{figure}

\noindent Furthermore, the continuous formulation of our generator provides great generalization capabililties even for points never seen during training.
This is shown by the high-resolution examples from the voxel-based approach in Figure~\ref{fig:generated_shapes} and a case study in Figure~\ref{fig:upscaling}.

\section{Conclusion}%
\label{sec:conclusion}

We presented two methods to produce continuous signed distance fields via generative adversarial modeling.
Further works include the study of more expressive point-processing discriminators, and the addition of regularization schemes to penalize invalid SDFs.
In addition, discriminator architectures based on differentiable Marching Cubes~\cite{deepmarchingcubes} or differentiable Sphere Tracing~\cite{differentiablespheretracing} could be examined.

\section*{Acknowledgments}

This work has been supported by the \emph{German Research Association (DFG)} within the Collaborative Research Center SFB 876, \emph{Providing Information by Resource-Constrained Analysis}, project A6.

\bibliographystyle{eg-alpha-doi}
\bibliography{egbibsample}

\newcommand{\etalchar}[1]{$^{#1}$}
\begin{thebibliography}{\uppercase{GPAM{\etalchar{*}}14}}

\bibitem[ADMG17]{pointcloud-gan}
\textsc{Achlioptas P., Diamanti O., Mitliagkas I., Guibas L.}:
\newblock Learning representations and generative models for {3D} point clouds.
\newblock \emph{CoRR} (2017).
\newblock \href {http://arxiv.org/abs/1707.02392} {\path{arXiv:1707.02392}}.

\bibitem[ASS{\etalchar{*}}18]{survey}
\textsc{Ahmed E., Saint A., Shabayek A., Cherenkova K., Das R., Gusev G.,
  Aouada D.}:
\newblock A survey on deep learning advances on different {3D} data
  representations.
\newblock \emph{CoRR} (2018).
\newblock \href {http://arxiv.org/abs/1808.01462} {\path{arXiv:1808.01462}}.

\bibitem[CFG{\etalchar{*}}]{shapenet}
\textsc{Chang A.~X., Funkhouser T., Guibas L., Hanrahan P., Huang Q., Li Z.,
  Savarese S., Savva M., Song S., Su H., Xiao J., Yi L., Yu F.}:
\newblock {ShapeNet}: An information-rich {3D} model repository.
\newblock \emph{CoRR}.
\newblock \href {http://arxiv.org/abs/1512.03012} {\path{arXiv:1512.03012}}.

\bibitem[CZ19]{implicit-fields}
\textsc{Chen Z., Zhang H.}:
\newblock Learning implicit fields for generative shape modeling.
\newblock In \emph{CVPR} (2019).

\bibitem[DQN17]{3dcnn-sdf}
\textsc{Dai A., Qi C.~R., Nie{\ss}ner M.}:
\newblock Shape completion using {3D}-encoder-predictor {CNNs} and shape
  synthesis.
\newblock In \emph{CVPR} (2017).

\bibitem[FLWM18]{splinecnn}
\textsc{Fey M., Lenssen J.~E., Weichert F., M{\"u}ller H.}:
\newblock {SplineCNN}: Fast geometric deep learning with continuous {B-Spline}
  kernels.
\newblock In \emph{CVPR} (2018).

\bibitem[GAA{\etalchar{*}}17]{wgan-gp}
\textsc{Gulrajani I., Ahmed F., Arjovsky M., Dumoulin V., Courville A.~C.}:
\newblock Improved training of {Wasserstein GANs}.
\newblock In \emph{NIPS} (2017).

\bibitem[GFK{\etalchar{*}}18]{atlasnet}
\textsc{Groueix T., Fisher M., Kim V.~G., Russell B.~C., Aubry M.}:
\newblock {AtlasNet}: A papier-m\^ach\'e approach to learning {3D} surface
  generation.
\newblock In \emph{CVPR} (2018).

\bibitem[GPAM{\etalchar{*}}14]{gan}
\textsc{Goodfellow I., Pouget-Abadie J., Mirza M., Xu B., Warde-Farley D.,
  Ozair S., Courville A., Bengio Y.}:
\newblock Generative adversarial nets.
\newblock In \emph{NIPS} (2014).

\bibitem[GYW{\etalchar{*}}19]{sdm-net}
\textsc{Gao L., Yang J., Wu T., Yuan Y.-J., Fu H., Lai Y.-K., Zhang H.}:
\newblock {SDM-NET}: Deep generative network for structured deformable mesh.
\newblock \emph{CoRR} (2019).
\newblock \href {http://arxiv.org/abs/1908.04520} {\path{arXiv:1908.04520}}.

\bibitem[Har96]{spheretracing}
\textsc{Hart J.~C.}:
\newblock Sphere tracing: A geometric method for the antialiased ray tracing of
  implicit surfaces.
\newblock \emph{The Visual Computer 12}, 10 (1996), 527--545.

\bibitem[HHF{\etalchar{*}}19]{meshcnn}
\textsc{Hanocka R., Hertz A., Fish N., Giryes R., Fleishman S., Cohen-Or D.}:
\newblock {MeshCNN}: A network with an edge.
\newblock In \emph{SIGGRAPH} (2019).

\bibitem[KALL18]{progressive-gan}
\textsc{Karras T., Aila T., Laine S., Lehtinen J.}:
\newblock Progressive growing of {GANs} for improved quality, stability, and
  variation.
\newblock In \emph{ICLR} (2018).

\bibitem[KH13]{psr}
\textsc{Kazhdan M., Hoppe H.}:
\newblock Screened poisson surface reconstruction.
\newblock \emph{ACM Transactions on Graphics 32}, 3 (2013), 29:1--29:13.

\bibitem[LC87]{marchingcubes}
\textsc{Lorensen W.~E., Cline H.~E.}:
\newblock Marching cubes: A high resolution {3D} surface construction
  algorithm.
\newblock In \emph{SIGGRAPH} (1987).

\bibitem[LDG18]{deepmarchingcubes}
\textsc{Liao Y., Donne S., Geiger A.}:
\newblock Deep marching cubes: Learning explicit surface representations.
\newblock In \emph{Proceedings of the IEEE Conference on Computer Vision and
  Pattern Recognition} (2018), pp.~2916--2925.

\bibitem[LW19]{meta-functions}
\textsc{Littwin G., Wolf L.}:
\newblock Deep meta functionals for shape representation.
\newblock In \emph{ICCV} (2019).

\bibitem[LZP{\etalchar{*}}19]{differentiablespheretracing}
\textsc{Liu S., Zhang Y., Peng S., Shi B., Pollefeys M., Cui Z.}:
\newblock Dist: Rendering deep implicit signed distance function with
  differentiable sphere tracing.
\newblock \emph{arXiv preprint} (2019).
\newblock \href {http://arxiv.org/abs/1911.13225} {\path{arXiv:1911.13225}}.

\bibitem[MON{\etalchar{*}}19]{occupancy-networks}
\textsc{Mescheder L., Oechsle M., Niemeyer M., Nowozin S., Geiger A.}:
\newblock Occupancy networks: Learning {3D} reconstruction in function space.
\newblock In \emph{CVPR} (2019).

\bibitem[PFS{\etalchar{*}}19]{deepsdf}
\textsc{Park J.~J., Florence P., Straub J., Newcombe R., Lovegrove S.}:
\newblock {DeepSDF}: Learning continuous signed distance functions for shape
  representation.
\newblock In \emph{CVPR} (2019).

\bibitem[QSMG17]{pointnet}
\textsc{Qi C.~R., Su H., Mo K., Guibas L.~J.}:
\newblock {PointNet}: Deep learning on point sets for {3D} classification and
  segmentation.
\newblock In \emph{CVPR} (2017).

\bibitem[QYSG17]{pointnet2}
\textsc{Qi C.~R., Yi L., Su H., Guibas L.~J.}:
\newblock {PointNet++}: Deep hierarchical feature learning on point sets in a
  metric space.
\newblock In \emph{NIPS} (2017).

\bibitem[RMC15]{dcgan}
\textsc{Radford A., Metz L., Chintala S.}:
\newblock Unsupervised representation learning with deep convolutional
  generative adversarial networks.
\newblock \emph{CoRR} (2015).
\newblock \href {http://arxiv.org/abs/1511.06434} {\path{arXiv:1511.06434}}.

\bibitem[SUHR17]{learning-geometry-images}
\textsc{Sinha A., Unmesh A., Huang Q., Ramani K.}:
\newblock {SurfNet}: Generating {3D} shape surfaces using deep residual
  networks.
\newblock In \emph{CVPR} (2017).

\bibitem[TDB17]{ogn}
\textsc{Tatarchenko M., Dosovitskiy A., Brox T.}:
\newblock Octree generating networks: Efficient convolutional architectures for
  high-resolution {3D} outputs.
\newblock In \emph{ICCV} (2017).

\bibitem[VFM19]{pointcloud-gan-iclr}
\textsc{Valsesia D., Fracastoro G., Magli E.}:
\newblock Learning localized generative models for 3d point clouds via graph
  convolution.
\newblock In \emph{ICLR} (2019).

\bibitem[WZL{\etalchar{*}}18]{mesh-deformation}
\textsc{Wang N., Zhang Y., Li Z., Fu Y., Liu W., Jiang Y.-G.}:
\newblock {Pixel2Mesh}: Generating {3D} mesh models from single {RGB} images.
\newblock In \emph{ECCV} (2018).

\bibitem[WZX{\etalchar{*}}16]{3dcnn}
\textsc{Wu J., Zhang C., Xue T., Freeman B., Tenenbaum J.}:
\newblock Learning a probabilistic latent space of object shapes via {3D}
  generative-adversarial modeling.
\newblock In \emph{NIPS} (2016).

\bibitem[YHCOZ18]{pointcloud-net}
\textsc{Yin K., Huang H., Cohen-Or D., Zhang H.}:
\newblock {P2P-NET}: Bidirectional point displacement net for shape transform.
\newblock \emph{ACM Transactions on Graphics 37}, 4 (2018), 152.

\end{thebibliography}

\end{document}